\documentclass[runningheads,a4paper]{llncs}

\usepackage{tikz}
\usepackage{fancyvrb}
\usepackage{listings}

\usepackage{caption}
\usepackage{subcaption}

\newcommand{\rosoclingo}{\textit{ROSoClingo}}
\newcommand{\turtlebot}{\emph{TurtleBot}}
\newcommand{\oclingo}{\textit{oClingo}}
\newcommand{\iclingo}{\textit{iClingo}}
\newcommand{\movebase}{\texttt{move\_base}}
\newcommand{\actionlib}{\emph{actionlib}}
\newcommand{\outrosoclingo}{\emph{out\_rosoclingo}}
\newcommand{\inrosoclingo}{\emph{in\_rosoclingo}}
\newcommand{\gazebo}{\emph{Gazebo}}
\newcommand{\deliver}{\texttt{deliver}}
\newcommand{\pickup}{\texttt{pickup}}
\newcommand{\comment}[1]{}

\begin{document}

\mainmatter  % start of an individual contribution

\title{\rosoclingo:\\ A ROS package for ASP-based robot control\thanks{This paper is also submitted at the Combined Planning Workshop RSS 2013.}}

\titlerunning{\rosoclingo}

\author{Benjamin Andres$^1$
\and Philipp Obermeier$^1$
\and Orkunt Sabuncu$^1$
\and Torsten Schaub$^1$
\and
David Rajaratnam$^2$ \\
\institute{$^1$University of Potsdam, $^2$University of New South Wales}
\email{$^1$\{bandres,phil,orkunt,torsten\}@cs.uni-potsdam.de \\ $^2$daver@cse.unsw.edu.au}
}
\authorrunning{Andres et al.}


%\email{\{bandres,phil,orkunt,torsten\}@cs.uni-potsdam.de}
%\url{http://www.springer.com/lncs}}

%\title[\rosoclingo]
%{\rosoclingo:\\ A ROS package for ASP-based robot control}
%\author[Andres et al.]{
%Benjamin Andres, Philipp Obermeier, Orkunt Sabuncu and Torsten Schaub %\\
%University of Potsdam \\
%\email{\{bandres,phil,orkunt,torsten\}@cs.uni-potsdam.de} \\
%\and
%David Rajaratnam \\
%University of New South Wales \\
%daver@cse.unsw.edu.au \\
%\email{daver@cse.unsw.edu.au}
%}

\maketitle

%\footnotetext{This paper is also submitted at the Combined Planning Workshop RSS 2013.}
\begin{abstract}
Knowledge representation and reasoning capacities are vital to cognitive robotics because they
provide higher level cognitive functions for reasoning about actions, environments, goals, perception, etc.
Although Answer Set Programming (ASP) is well suited for modelling such functions, 
there was so far no seamless way to use ASP in a robotic environment.
We address this shortcoming and show how a recently developed reactive ASP system can be harnessed to
provide appropriate reasoning capacities within a robotic system.
To be more precise,
we furnish a package integrating the reactive ASP solver \oclingo\ with the popular open-source
robotic middleware ROS.
The resulting system, \rosoclingo, provides a generic way by which an ASP program can be
used to control the behaviour of a robot and to respond to the results of the robot's actions.
\end{abstract}

%%% Local Variables: 
%%% mode: latex
%%% TeX-master: "paper"
%%% End: 

\section{Introduction}

Knowledge representation and reasoning capacities are vital to cognitive robotics because they
provide higher level cognitive functions for reasoning about actions, environments, goals, perception, etc.
Although Answer Set Programming (ASP) is well suited for modelling such functions, 
there was so far no seamless way to use ASP in a robotic environment.
This is because ASP solvers were designed as one-shot problem solvers and thus lacked any reactive
capacities.
So, for instance, each time new information arrived, the solving process had to be re-started from scratch.

In what follows, we address this shortcoming and show how a recently developed \emph{reactive} ASP
system \cite{gegrkasc11a,gegrkaobsasc12b} can be harnessed to provide knowledge representation and
reasoning capacities within a robotic system.
This is possible because such systems allow for incorporating online information into operative ASP
solving processes.
We accomplish this by integrating our ASP approach into the popular open-source middleware
ROS\footnote{\texttt{http://www.ros.org}}
(Robot Operating System;~\cite{quigley09}) 
which has become a de facto standard in robotics over the last years.
As such, ROS provides hardware abstraction and tools supporting the development of robot
applications.

To be more precise,
we furnish a ROS package integrating the reactive ASP solver \oclingo\ with the popular open-source
ROS robotic middleware.
The resulting system, called \rosoclingo, provides a generic method by which an ASP program can be
used to control the behaviour of a robot and to respond to the results of the robot's actions.
In this way,
the \rosoclingo\ package plays the central role in fulfilling the need for high-level knowledge
representation and reasoning in cognitive robotics by making details of integrating a highly capable
reasoning framework within a ROS based system transparent for developers.
In this regard, \rosoclingo\ can also function as an important component for
bridging the gap between high-level task planning and low-level motion planning.
In what follows,
we provide the architecture and basic functioning of the \rosoclingo\ system.
And we illustrate its operation via a case-study conducted with the ROS-based
\turtlebot \footnote{\texttt{http://turtlebot.com}}
simulation in a \emph{Gazebo}\footnote{\texttt{http://gazebosim.org}} simulation of an office floor.
The resulting work is publicly available\footnote{\texttt{http://www.cs.uni-potsdam.de/wv/ROSoClingo/index.html}} and we are committed to
submit the \rosoclingo\ package to the public ROS repository.

The Golog programming language \cite{DBLP:journals/jlp/LevesqueRLLS97}
is one of the most widely known approaches to the development of a
declarative agent reasoning language. With a formal semantics based on
the Situation Calculus \cite{McCHay69} it allows for the specification
of high-level agent behaviours for agents acting within dynamically
changing environments. The potential power of this approach was first
shown on a real robot with the implementation of the Golex system
\cite{DBLP:conf/ki/HahnelBL98}. Golex extended Golog with execution
monitoring functionality to monitor and ensure the successful
execution of the primitive Golog actions.

With the success of Golog, further work has focused on extending ASP
with some of the expressive constructs found in Golog
\cite{DBLP:conf/asp/SonBM01}, thus allowing the powerful search
capabilities of modern reasoners to be combined with the programming
ease of Golog.  The development of \rosoclingo\ can therefore be
understood in the context of allowing Golog, and other, ASP extensions
for agent reasoning to be directly applied to the high-level control
of ROS based robots.

Finally, we list some related work which utilize ASP or other declarative formalisms in 
cognitive robotics.
The work in \cite{chjijijiwa09a,chjijijiwaxi10a} uses ASP for 
representing knowledge via a natural language based human robot interface. 
Additionally, ASP is used for high level task planning.
In \cite{akererpa11a,erakpa12a} action language formalism and ASP are used to
plan and coordinate multiple robots for fulfilling an overall task.
They have also integrated task and motion planning with external calls 
from action formalism to geometric reasoning modules \cite{erhapapaur11a}.
All
%resulting implementations rely on one-shot ASP solvers and thus lack any reactive capacities.
%Hence, all 
these works can naturally and highly benefit from the usage of \rosoclingo.
Having stated that,
\rosoclingo\ can be basically used in any autonomous robotics system in which 
high-level reasoning tasks are essential and steep initial integration difficulties
are desired to be avoided.

%%% Local Variables: 
%%% mode: latex
%%% TeX-master: "paper"
%%% End: 

\section{Robot Operating System}
\label{sec:ros-background}

ROS provides a middleware for robotic applications
\cite{quigley09}. At its most basic level this consists of a
loosely-coupled communication framework for sending \emph{messages}
between processes. ROS defines a host and language independent TCP/IP
protocol for exchanging messages, thus allowing these processes to be
written in a variety of programming languages and to be distributed
across multiple host computers.

ROS standardises methods and structures for organising software into
\emph{packages}. Packages can contain a variety of components, from
definitions of message formats through to libraries and executable
programs. Executable programs that integrate into the ROS framework
are instantiated as special processes known as \emph{nodes}.

\begin{figure}[t]
\centering
\includegraphics[width=0.5\textwidth]{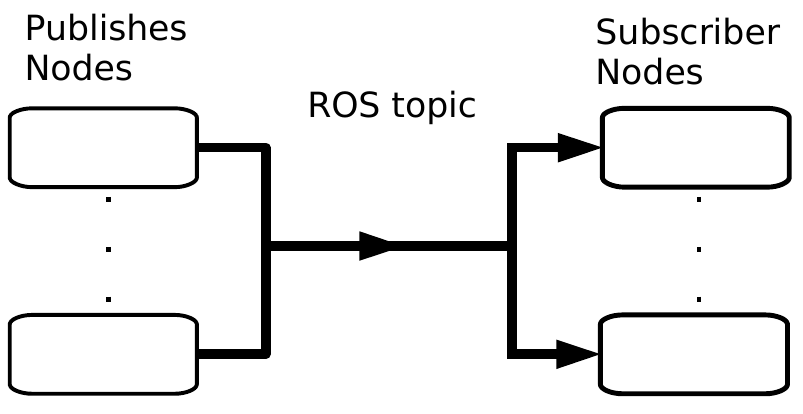}
\caption{\label{fig:ros_topics}ROS topics provide many-to-many asynchronous message passing.}
\end{figure}

There are two basic mechanisms for communications between nodes. The
\emph{publisher-subscriber} mechanism provides for asynchronous
co\-mu\-ni\-ca\-tions where\-by multiple nodes can broadcast messages on a
named communication channel (known as a \emph{topic}), that are
in-turn listened to by multiple subscribers
(Figure~\ref{fig:ros_topics}). Alternatively, ROS \emph{services}
provide for synchronous communication via a remote procedure call
(RPC) mechanism whereby one node can call a service provided by
another node.

%\begin{figure}[Ht!]

All messages in ROS are strongly-typed, and all communications using
topics or services must use these types. ROS provides a number of
primitive data types (e.g., \texttt{bool}, \texttt{int32},
\texttt{float32}, \texttt{string}) as well as a list operator. These
can be combined to produce arbitrarily complex types in a similar
manner to \texttt{structs} in C and C++.  These complex data types are
defined as part of the ROS package structure. To ease development and
code maintenance ROS package names inherently correspond to namespaces
of the same name and therefore the complex data types are always
defined with respect to a namespace. A typical ROS system defines a
number of common namespaces (e.g., \texttt{std\_msgs},
\texttt{geometry\_msgs}) and data types (e.g.,
\texttt{geometry\_msgs/Pose}).

\begin{figure}[Ht!]
%\begin{figure}
\centering
\includegraphics[width=0.6\textwidth]{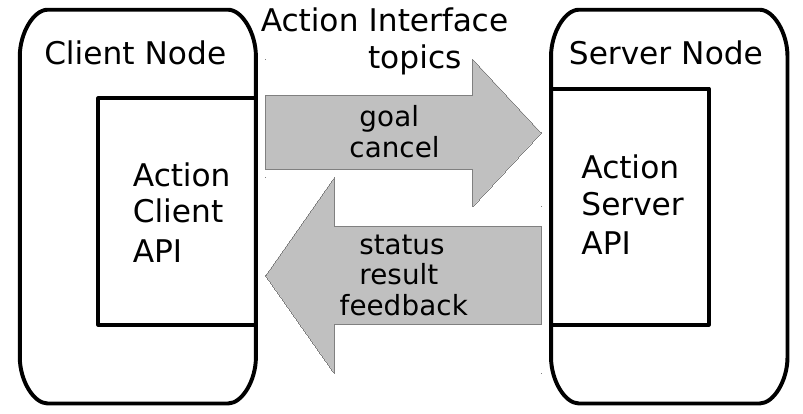}
\caption{\label{fig:ros_actions}ROS actions provide preemptible client-server communications.}
\end{figure}

While ROS services allow for a simple RPC mechanism, they are not
suitable for more complex behaviours, such as situations where a task
may take place over an extended time frame, may be preempted, and may
require feedback throughout its life-cycle. ROS provides for such
complex behaviour through the \actionlib\ framework
(Figure~\ref{fig:ros_actions}). This framework allows a client-server
interface to be defined whereby an \emph{action client} is able to set
and cancel goals on an \emph{action server}. The action server in turn
executes the goal and provides constant feedback and progress of its
attempts to fulfil the goal. While implemented using ROS topics as a
message transport mechanism, each \emph{action interface} defines a
high-level API for client-server interaction.\footnote{For more
  extensive information on programming with the ROS \actionlib\
  framework the interested reader is referred to
  \texttt{http://www.ros.org/wiki/actionlib}}

As a prototypical example of a ROS action, the \movebase\ package
implements an action interface that provides path-planning and robot
control functionality for moving a robot around an environment. We now
briefly outline this package, to provide both a sense of how the
\actionlib\ framework works as well as to provide details of an
important module that we shall discuss later.

\begin{figure}[Ht!]
%\begin{figure}
\centering
\includegraphics[width=0.8\textwidth]{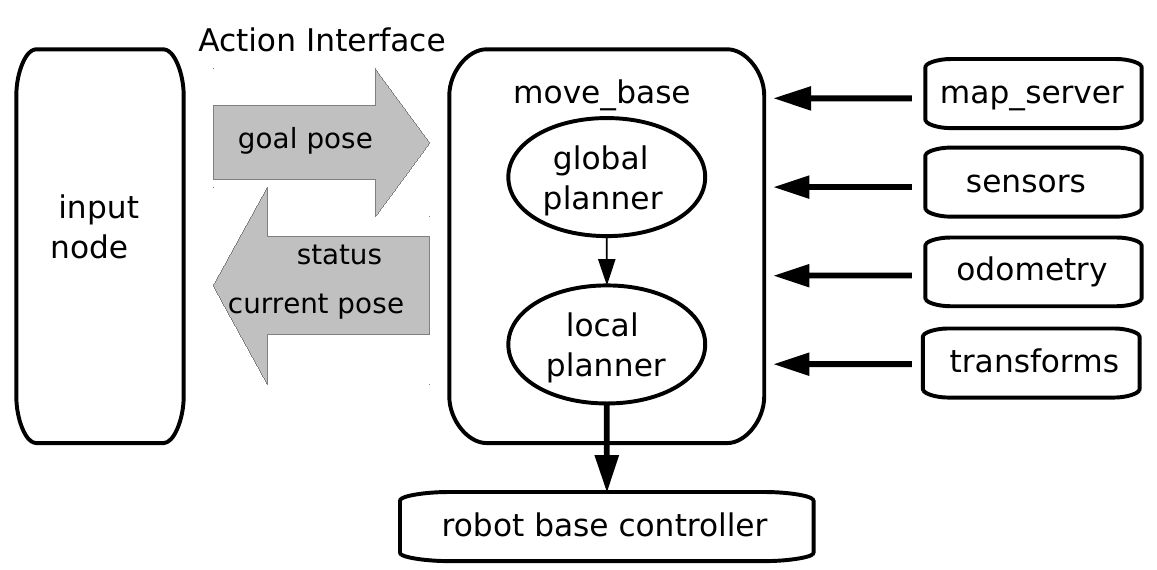}
\caption{\label{fig:ros_move_base}ROS \movebase\ provide an action interface for robot movement.}
\end{figure}

The \movebase\ package (Figure~\ref{fig:ros_move_base}) provides a
highly configurable ROS node that is an essential component of the ROS
\emph{navigation stack}. The system requires a number of data sources,
such as laser sensor data, localisation information, and map
information. The laser sensor data is used to perform basic obstacle
avoidance, while localisation information allows the robot to know
where it is located within the map. A map consists of a 2-D
\emph{occupancy grid} that indicates whether a point on the grid is
occupied or free. Robot navigation within the node takes place at two
distinct levels. \emph{Global planning} calculates the route from the
robot's current location to a goal location, while \emph{local
  planning} provides for movement towards a general direction while
allowing path flexibility to avoid obstacles. A range of different
global and local planning algorithms are supported through a plugin
architecture.

%\begin{figure}
%\centering
%\parbox{6cm}{
%  \includegraphics[width=0.4\textwidth]{Figures/ros_actions.pdf}
%  \caption{\label{fig:ros_actions}ROS actions provide preemptible client-server communications.}
%\label{fig:2figsA}}
%\qquad
%\begin{minipage}{6cm}
%  \includegraphics[width=1.0\textwidth]{Figures/ros_move_base.pdf}
%  \caption{\label{fig:ros_move_base}ROS \movebase\ provide an action interface for robot movement.}
%\label{fig:2figsB}
%\end{minipage}
%\end{figure}

\comment{
% THIS DISCUSSION IS MAYBE A BIT OF A SIDE-TRACK AND NOT REALLY NECESSARY
As might be expected, localisation and coordinate systems play an
important role in any robotic application. A central aspect of the ROS
approach to this problem is the maintenance and broadcasting of a tree
of \emph{transformations} between different frames of reference. This
allows individual nodes to use the frame of reference that is most
appropriate to its needs while still allowing for coordinates to be
transformed to some other frame of reference when necessary. For
example the location of an object in a home environment is most
usually specified relative to the map of the home, while moving a
robot arm is appropriately performed relative to the robot
itself. Hence making the robot arm pick up the object requires
translating between these two frames of reference.  }

A navigation goal is specified in terms of a robot destination
\emph{pose} (i.e., position and orientation). When a goal is sent to
the \movebase\ server it computes a path to that goal location and
then successively generates the movement commands for the robot base
controller. If at some point the robot is unable to proceed with its
plan, for example due to a door being blocked, then the server will
undertake recovery behaviour and will re-plan accordingly. If the
recovery fails then the task will be aborted. Throughout this process
the server provides constant feedback as to the current location of
the robot, as well as the status of the navigation process, for
example that the task has been aborted. Goals are preemptible, so that
a robot navigating towards some location will give up that goal if it
is given a new destination goal.

The action interface provided by the \movebase\ package is arguably
the most important ROS action service for mobile robots. Furthermore,
with a complex set of features, such as the possibility of failure, it
serves to highlight the potential complexity in trying to integrate
logical reasoning with a real robotic systems. Consequently, the
\movebase\ package forms much of the integration work that is outlined
in the rest of this paper.

%%% Local Variables: 
%%% mode: latex
%%% TeX-master: "paper"
%%% End: 

\subsection{\oclingo}

A classical ASP system, such as gringo/clasp, is designed to solve
problems in an one-shot procedure: it takes a problem encoding as
input, computes the answer sets and terminates afterwards.
Since this approach does not fully embrace the needs of modern
dynamic domains, such as robotics, a reactive ASP solver, \oclingo{},
was developed that additionally takes external data streams into
account.
Such a stream is represented there by an \emph{online progression}, a sequence of events, i.e., data updates and inquiries, given in the form of ASP ground facts and integrity constraints. 
The general problem itself is described by a \emph{reactive logic
  program}, an ASP program \if(set of normal logic programming clauses)\fi
that is partitioned into three parts: a \emph{base} part\ \if\(B\)\fi describing
static knowledge, and an \emph{incremental} as well as a \emph{volatile}
part \if\(P\) and \(Q\)\fi which both contain rule schemata based on a discrete time
(integer) parameter\if \(t\)\fi. The role of the incremental
part is to symbolize accumulated knowledge over increasing time,
whereas the volatile part only holds information that specifically
concerns the current point in time.
All in all, a reactive logic program formulates the persistent
knowledge and, thus, acts as the offline counterpart to an online
progression. We will further illustrate these concepts by examples in
Section \ref{sec:casestudy}.

\begin{figure}[Ht!]
\centering
\includegraphics[width=0.8\textwidth]{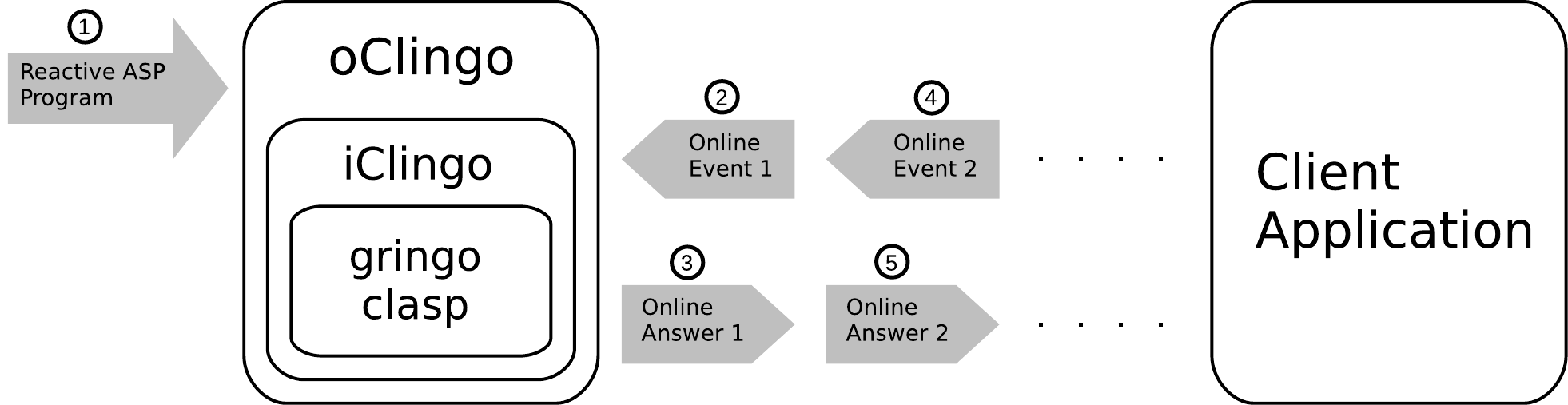}
\caption{\label{fig:oclingo} Basic workflow between oClingo and a client application.}
\end{figure}

% % Technical aspects

%
% Technically, the \oclingo{} system is initialized with a reactive
% logic program as input (Figure~\ref{fig:oclingo}). 
%
Technically, \oclingo{} relies and expands on \iclingo{}~\cite{gekakaosscth08a}, an
incremental extension of gringo/clasp, as well as on gringo/clasp itself.
That is, the \oclingo{} system is initialized with a reactive logic
program as input (Figure~\ref{fig:oclingo}).
After that, a client application can connect to \oclingo{} and stream
data formatted as an online progression.
%
% For each incoming stream update \oclingo{} computes all answers,
% returns them to the client and subsequently waits for the next input.
%
For each incoming event \oclingo{} instructs gringo to add the
contained clauses to the existing ground program.
This addition of new information is either temporarily (valid only for
the current event) or persistent depending on the lifetime annotation
provided by the event for each clause.
Subsequently, \oclingo{} invokes \iclingo{} to compute all answers for the current
ground program, returns them to the client and waits afterwards for
new events.
The \oclingo{} system only terminates if explicitly requested by the client.
\if0
\color{blue}
\begin{itemize}
\item oclingo 
  \begin{itemize}
  \item instructs grounder gringo to add to the existing ground program the new input
  \item instruct iclingo to ground the incremental ground program to level requested by the us
  \item instructs iclingo to repeatedly run incremental solving cycles until the next answer is found
  \end{itemize}
\end{itemize}
\color{black}
\fi
\if0
\color{blue}
\begin{itemize}
\item General
  \begin{itemize}
  \item A classical ASP system, such as gringo/clasp, is designed to
    solve problems in a one-shot procedure: it takes a problem
    encoding as input, computes the answer sets and terminates
    afterwards.
  \item Since this approach does not fully embrace the needs of modern
    dynamic domains, such as robotics, a reactive ASP solver,
    \oclingo, was developed that additionally takes external data
    streams into account.
  \end{itemize}
\item Syntax, Semantics, Technical Terminology
  \begin{itemize}
  \item (Time-decaying) incremental logic programs.
    \begin{itemize}
    \item Incremental logic program
      \begin{itemize}
      \item incrementally growing program until a minimal answer/horizon can be found
      \end{itemize}
    \item Time-decaying logic program
      \begin{itemize}
      \item technical \texttt{\#volatile}
      \end{itemize}
    \item Time-decaying online progression
      \begin{itemize}
        \item controller, stream
        \item \texttt{\#step. ... \#endstep.} events
      \end{itemize}
    \end{itemize}
  \end{itemize}
\end{itemize}
\color{black}
\fi

\section{\rosoclingo }

\begin{figure}[Ht!]
\centering
    \includegraphics[width=0.75\textwidth]{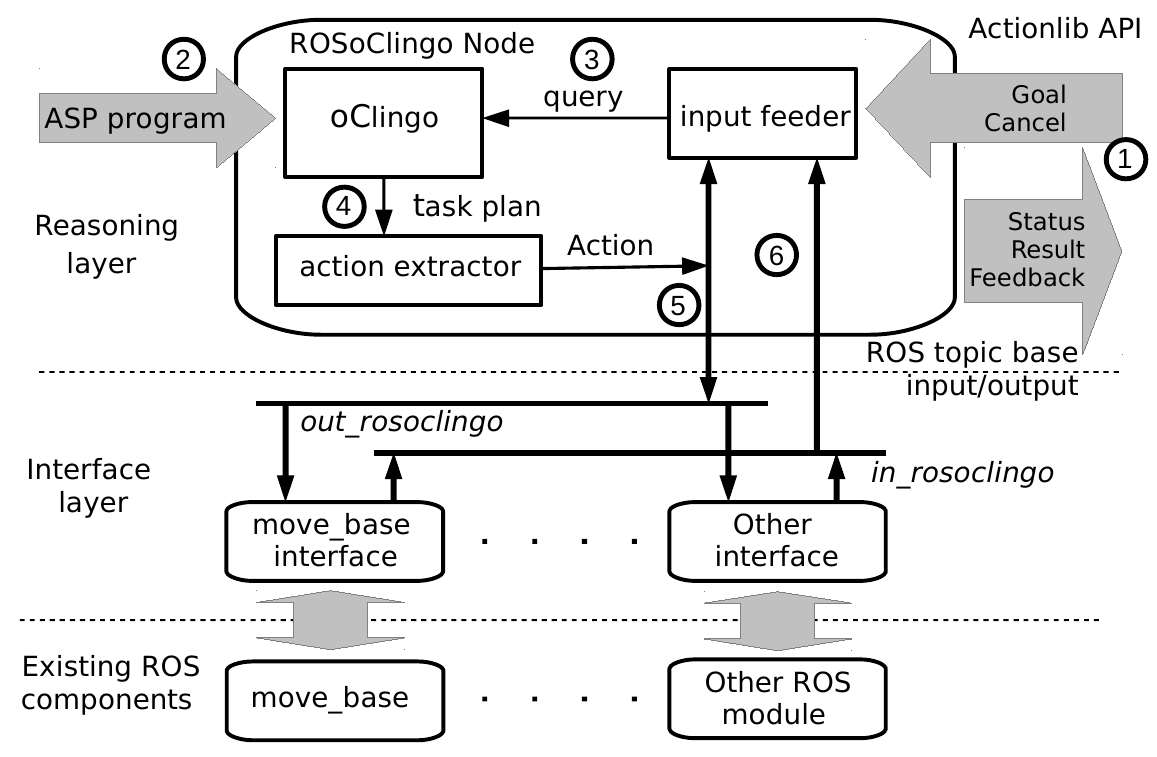}
\caption{\label{fig:workflow}The general architecture and main workflow of \rosoclingo.}
\end{figure}

In this section we describe the general architecture and functionality
of the \rosoclingo\ package.  With the help of reactive ASP,
\rosoclingo\ provides a way of handling high-level knowledge
representation and reasoning tasks occurring in autonomous robots
running the ROS software.

Consider a task planning problem, a task that any autonomous robot
should be capable of performing. For instance, a robot is given a goal
of moving from the kitchen of a house to the living room in order to
serve food.
Even if the robot has many individual behaviours, like moving from one
point to another or holding food with the help of respective ROS
packages, computing a complete task plan
%, where a specific path among various possible ones is selected,
requires high-level knowledge representation and reasoning
capabilities.
There might even be more than one possible path to the living room,
which may require more elaborate planning and execution.  The robot
might, for instance, need to first go to the hallway and then to the
living room in case the alternative path is blocked by an obstacle.
The robotics developer can encode such planning tasks in reactive ASP
keeping only the interface requirements of the underlying behaviour
nodes in mind and avoiding implementation details of their
functionality (motion planning for example).  Then, using the
resulting ASP encoding with the \rosoclingo\ package the developer can readily
integrate task planning while details of controlling and integrating
\oclingo\ within the ROS middleware becomes transparent.

Figure \ref{fig:workflow} depicts the main components and workflow of
the \rosoclingo\ system. It consists of a three layered
architecture. The first layer consists of the core
\rosoclingo\ component and the definition of an \actionlib\ API. This
API allows other components to use the services provided by the
\rosoclingo\ node. The package also defines the message structure for
communication between the core \rosoclingo\ node and the various nodes
of the interface layer. The interface layer, on the other hand,
provides the data translations between what is required by the
\rosoclingo\ node and any ROS components for which it needs to
integrate. This architecture provides for a clean separation of
duties, with the well-defined abstract reasoning tasks handled by the
core node and the integration details handled by the interface nodes.

\begin{figure}[Ht!]
%\begin{figure*}[Ht!]
\centering
\begin{tabular}{|lll|}
\cline{1-3}
%\multicolumn{3}{|l|}{\bf Keyords}\\
% & & \vspace{-1.5mm}\\
& \texttt{\_request(Goal,C)} & Specifying a Goal request at cycle \texttt{C}. \\ %Cancellations are send in the same way.\\
& & Cancellations are send in the same way. \\
& \texttt{\_action\_lib(A,P,C)} & Commanding \actionlib\ \texttt{A} to run with parameters \texttt{P} at cycle \texttt{C}.\\
& \texttt{\_return(A,V,C)} & Specifying the return value \texttt{V} of \actionlib\ \texttt{A} run in cycle \texttt{C}.\\
\cline{1-3}
\end{tabular}
\caption{\label{fig:keywords} Keywords used for communicating between \rosoclingo\ and \oclingo.}
\end{figure}

\subsection{The \rosoclingo\ Core}

The main \rosoclingo\ node is composed of
%Main components of \rosoclingo\ are 
the reactive answer set solver \oclingo, an action extractor, and an input
feeder. Through its \actionlib\ API, it can receive goal and
cancellation requests as well as send result, feedback, and status
messages to a client node (marked by 1 in the figure).
The reactive ASP program, encoding the high-level task planning problem, is given to the
\rosoclingo\ node at system initialization (marked by 2).
During initialization, \rosoclingo\ sets the current logical time point to 1.
This time point is incremented at the end of each cycle.

A cycle of \rosoclingo's workflow may start with a goal for the robot arriving
% Goals for the robot can arrive 
via the \actionlib\ interface (marked by 1).
For instance, commanding the robot to go to the living room can be a goal request.
This request is transformed into an input stream update before feeding to \oclingo\ (marked by
3) by the input feeder.
\oclingo\ receives the goal as a stream update and searches for an answer set representing
a task plan for the robot to follow.
% This plan is list of actions ordered by logical time points.
Each action of the plan, in principle, should be executed by a respective ROS node.
For instance, an action of moving to the door connecting kitchen and hallway can be
executed by the \movebase\ ROS action node.
The keywords of Figure~\ref{fig:keywords} allow for a communication protocol between \rosoclingo\ and \oclingo.
The action extractor takes the action at the current logical time point, 
prepares it as a goal request (marked by 4) and
sends it to be executed by the respective ROS node (marked by 5).
The result of the execution is received by the input feeder component of 
the \rosoclingo\ node (marked by 6).
The communication between \rosoclingo\ and other ROS nodes is detailed in 
Section \ref{sec:rosoclingoandothernodes}.
The result is processed and transformed into a new input stream update for \oclingo, 
which completes the current and initiates a new cycle of \rosoclingo.
% According to type of the result of an action execution different input queries
% can be fed to \oclingo.
%
The (un)successful result may generate new knowledge for the robot about the world
(for example, the fact that a doorway is blocked or a new object is sensed).
Additionally, the next input update includes a fact so that 
the action executed is committed 
by \oclingo\ during further searches for a plan.

%Note that the \rosoclingo\ node exhibits preemptive behaviour.
%When a new goal arrives during the execution of a current one, 
%it cancels the current goal and processes the new goal at the next cycle.
%The status of each goal is also tracked by \rosoclingo.

Note that the \rosoclingo\ package supports multiple goal requests at a time.
Each time a new goal received,
the input feeder appends the goal to a list and feeds it to \oclingo\ in the next cycle.
The status of each goal is also tracked by \rosoclingo.

\subsection{Integrating with Existing ROS Components}
\label{sec:rosoclingoandothernodes}

The core \rosoclingo\ node needs to issue commands to, and receive
feedback from, existing robotics components. The complexity of this
interaction is handled by the nodes at the interface layer
(Figure~\ref{fig:workflow}). Unlike the components of the reasoning
layer it is, unfortunately, not possible to define a single ROS
interface to capture all interactions that may need to take
place. Firstly, there will need to be data type conversions between
the individual modules. For example, the \movebase\ node expects a
robot \emph{pose} as its goal, while an action to move a robot arm
might require a more complex goal structure consisting of a set of
joint-trajectories. Turning ROS messages into a suitable set of \oclingo\
statements will therefore require data type conversions that are
specific for each action or service type.

A second complicating issue is that the level of abstraction of a ROS
action may not be at the appropriate level required by the ASP
program. For example, the \emph{pose} goal for moving a robot consists
of a Cartesian coordinate and orientation. However, it is unlikely
that one would want a logical reasoner to have to reason about
Cartesian coordinates. Instead one would hope to reason about abstract
locations and the relationship between these locations; for example
that the robot should navigate from the kitchen to the bedroom via the
hallway. Furthermore, the desired orientation of the robot when it
arrives in the bedroom may not be something that is of interest to the
reasoner.

While it is not possible to provide a single generic interface to all
ROS components, it is however possible to outline a common pattern for
such integration. The rest of this section outlines the integration of
\rosoclingo\ with ROS actions, and in particular the \movebase\ action
outlined in Section~\ref{sec:ros-background}. ROS actions typically
encapsulate the high-level behaviour and functionality of a robot, and
are therefore the most natural level at which a high-level robot
controller would expect to communicate with the rest of the robotic
system. Furthermore, they are arguably the most complex components of
a ROS system. Consequently, showing how \rosoclingo\ integrates with
existing ROS actions encapsulates all the complexity that one would
expect of integration with any other ROS component.

For each existing ROS component that needs to be integrated with
\rosoclingo\ there will need to be a corresponding interface
component. In some cases interface components can be combined into a
single ROS node to communicate with multiple lower-level ROS nodes,
but in general one can imagine a mostly one-to-one correspondence
between nodes of these two layers.

An important consequence of our architecture is that every interface
node needs to read every message that is published by the
\rosoclingo\ node on the output topic. It is therefore important that
the message format for the \outrosoclingo\ topic allows the
interface components to easily parse the messages and discard those
messages that are intended for a different component. 

The inputs to, and outputs from, a running \oclingo\ reasoner consist
of sets of facts. It is therefore the role of the ROS interface layer
to perform any data conversions between the ASP world of facts and the
low-level ROS commands. We adopt a straightforward message type (named
\texttt{rosoclingo/InterfaceIO}) to facilitate this process. This type
consists of an interface name and a list of text formatted facts.

%\begin{lstlisting}[xleftmargin=10mm]
%# ROSoclingo interface IO message.
%#
%# The interface name can be used to discard non-relevant 
%# messages when sent from ROSoclingo to an interface component.
%string interface_name
%string[] facts
%\end{lstlisting}

%\begin{Verbatim}[xleftmargin=10mm]
%# ROSoclingo interface IO message.
%#
%# The interface name can be used to discard non-relevant 
%# messages when sent from ROSoclingo to an interface component.
%string interface_name
%string[] facts
%\end{Verbatim}

When a message is sent from \rosoclingo\ to the interface layer,
individual interface components can quickly parse the interface name
to determine the intended recipient and discard non-relevant
messages. On the other hand, when sent from an interface component to
the \rosoclingo\ node, the interface name indicates the origin of the
fact, which may be useful, even if only as a debugging aid.

For the sake of simplicity and presentation it is useful to make some
assumptions, in showing how the system integrates with the
\movebase\ action. A common assumption in robotic applications is to
identify tagged points with an abstract location. For example some
coordinate location specifying a point in a bedroom, say
\texttt{(10.5,11.2)}, will be associated with the label ``bedroom''.
Navigation can then take place with reference to the tagged
locations. This technique works for a broad range of behaviours, such
as sending the robot to specific locations.

\section{Case Study}
\label{sec:casestudy}
%For demontration purposes we simulated a mailbot delivery problem unsing ROSoClingo with ros fuerte.

We demonstrate the application of our \rosoclingo\ package on 
a mail delivery scenario running ROS software.
The scenario consists of a robot, whose task is to pick up 
and deliver mail packages exchanged among offices \cite{thielscher04a}.
Whenever a mail delivery request is received, 
the robot has to go to the office requesting the delivery,
pick up the mail package, 
and go to the destination office in order to do the delivery.
In this scenario the robot is able to carry up to three packages and handle multiple requests at a time.
Delivery requests can also be cancelled during task execution.
If a request is cancelled 
when the package has already been picked up,
%regarding a package that has already been picked up,
it is delivered back to its origin for disposal.
The task has a highly dynamic nature and requires reasoning capacity
for detailed planning.

The robot we use is a \turtlebot,
which is well supported within the ROS community and commonly used for small delivery tasks.
Offering a mobile platform with an integrated Microsoft Kinect as a three dimensional sensor.
Our office building is provided by \gazebo, a simulator supported by ROS,
able to realistically simulate three dimensional environments.
This allows us to run the \turtlebot\ in a controlled, yet physically plausible environment,
while avoiding all too common problems associated with hardware,
e.g. short battery life, defunct components, etc.

% The assignment of the mail delivery robot is to transport packages in an office building.
% The robot is able to receive package delivery requests and the their
% cancellation via a wireless connection with the offices.
% Upon receiving a request the robot has to move to the sending office,
% accept the package,
% move to its destination and
% deliver the package.
% If a request is cancelled the robot ignores that request from then on if the respective package is
% not accepted, yet.
% Otherwise, the package is delivered to the nearest office for disposal.
% While there are no pending requests the robot is idle.
% The robot in question, a turtlebot, is able to handle several requests and transport up to three
% packages at a time.

\begin{figure}[Ht!]
\centering
    \includegraphics[width=0.8\textwidth]{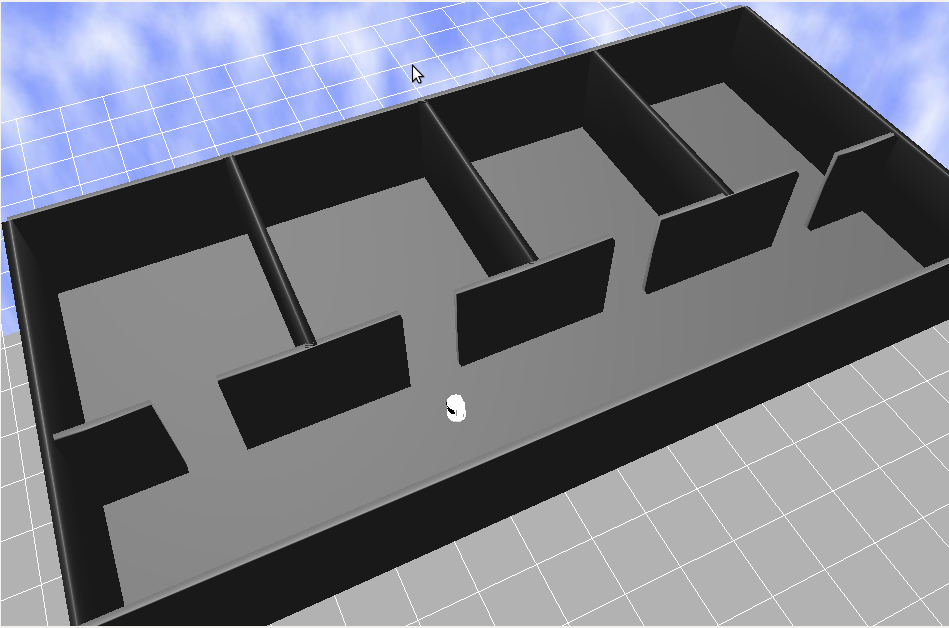}
\caption{\label{fig:gazebo_example} The mailbot simulation in progress as seen in \gazebo. 
%The offices are \texttt{office1} $\cdots$ \texttt{office4} appearing from left to right.
}
\end{figure}

We use the ROS \movebase\ action library for robot movement among offices.
For picking up and delivering packages, 
two dedicated action libraries are used (\pickup\ and \deliver ).
Figure \ref{fig:gazebo_example} shows the \gazebo\ environment used in our case study.
It consists of 4 consecutive offices on one floor;
\texttt{office1} to \texttt{office4} appearing from left to right.

% The move\_base action library was used for movement,
% but for accepting and delivering the packages two special purpose action libraries were created.
% For accepting a package the robot moves to the door of the office 
% and signals the employees inside to place the package on its tray and confirm the action by
% pressing a button.
% The delivery action library was fashioned analogical.

% As a minimalistic example the simulation environment consists of four offices all connected by one
% floor,
% as shown in figure \ref{}.
% Here, the robot is able to reach every office, their position labelled as \texttt{office1} to
% \texttt{office4}, by moving up and down the floor.

\begin{table}
 \caption{Task plans returned by \oclingo\ for varying requests.}
% \resizebox{0.65\columnwidth}{!}{
 \resizebox{\columnwidth}{!}
{
 \centering\scriptsize\tt
%\begin{tabular}{|r|@{\hspace{-2pt}}l@{}|@{\hspace{-2pt}}l@{}|@{\hspace{-2pt}}l@{}|}
\begin{tabular}{|r|l|l|l|}
%\begin{tabular}{|r|@{}l@{}|@{}l@{}|@{}l@{}|}
%\begin{tabular}{|r|l|l|l|}
\cline{1-4}
           plan & ~1 & ~2 & ~3 \\
\cline{1-4}
           step &                                  &                                       &                                       \\
1               & \_action(move\_base,office2,1)   & \_action(move\_base,office2,1)        & \_action(move\_base,office2,1)   \\
2               & \_action(move\_base,office3,2)   & \_action(move\_base,office3,2)        & \_action(move\_base,office3,2)   \\
3               & \_action(pickup,1,3)             &                                       &                                       \\
4               & \_action(move\_base,office2,4)   &                                       & \_action(pickup,2,4)   \\
5               & \_action(deliver,1,5)            &                                       & \_action(move\_base,office4,5)   \\
6               & 	    		           &                                       & \_action(deliver,2,6)            \\
\cline{1-4}
\end{tabular}
 }
 \label{table:example}
\end{table}

In our scenario the robot stands in front of \texttt{office1}
and after some time receives a request to deliver
a package from \texttt{office3} to \texttt{office2}.
This request is later cancelled and a new request to deliver a package from \texttt{office3} to \texttt{office4} is issued.

Since there are no pending requests just after the start of the simulation,
\oclingo\ returns an empty task plan.
This results in \rosoclingo\ awaiting a new request to be issued.
When the first request is received, the input feeder transforms said request into an input stream update for \oclingo\ : \\
\begin{lstlisting}[xleftmargin=5mm]
#step 1.
_request(goal(office3,office2,1),1).
#endstep.
\end{lstlisting}
The input stream update is enclosed in \verb|#step 1.| and \verb|#endstep.| directives.
The integer number following the \verb|#step| represents the cycle the request is send to \oclingo.
%
% With ``\verb|#step 1.|`` identifying the start of an input stream update and the cycle the request
% is send to \oclingo.
The keyword predicate \verb|_request| identifies the transformed goal request issued to
\rosoclingo\
with its first parameter and the cycle the request becomes active with its second.
The parameters of \verb|goal| state the sending office, 
the destination office, and an unique package identifier, 
in that order.
% ``\verb|#endstep.|`` closes the input stream update.

\oclingo\ now adapts the task plan to ensure the execution of the request as shown in Table \ref{table:example} under plan 1.
In more detail the plan involves to use the \texttt{move\_base} action library to move the
robot from \texttt{office1} over \texttt{office2} to \texttt{office3} at the cycles 1 and 2, respectively.
Then, the robot shall use the \texttt{pickup} action library to pick up the package in cycle 3
and move back to \texttt{office2} in cycle 4.
Lastly, the package is to be handed over by means of the \texttt{deliver} action library in cycle 5.
Note that both the \texttt{pickup} and the \texttt{deliver} action require the package identifier as parameter.

\rosoclingo's action extractor takes the task plan outputted as
an answer set by \oclingo\ and
publishes the action planned for the current (first) cycle on the
\outrosoclingo\ topic as a \texttt{rosoclingo/InterfaceIO} message.
The \texttt{move\_base} interface reacts to the message,
transforms the label \texttt{office2} into a coordinate location
and sends the result of the action back to
\rosoclingo\ via the \inrosoclingo\ topic.

In the next cycle the result is feed back into \oclingo\ using the keyword predicate
\verb|_return|.
Assuming \texttt{move\_base} was successful,
the input feeder generates the following input stream update for the second cycle:
\begin{lstlisting}[xleftmargin=5mm]
#step 2.
:- not _action(move_base,office2,1).
_return(move_base,office2,1).
#endstep.
\end{lstlisting}
The integrity constraint after the input stream header enforces \oclingo\ to include the action just taken into future action plans.
Otherwise, \oclingo\ might abolish actions taken in the past in order to minimize the task plan.
The rest of the cycle runs analogous to the first cycle shown. 
At sometime between the second and third cycle \rosoclingo\ receives the cancellation of the first request:
\begin{lstlisting}[xleftmargin=5mm]
#step 3.
:- not _action(move_base,office3,2).
_request(cancel(1),3).
_return(move_base,office3,2).
#endstep.
\end{lstlisting}
Here \verb|cancel(1)| identifies the delivery to be cancelled via its package identifier.
This forces \oclingo\ to change the task plan to the one presented under 
plan 2 in Table \ref{table:example}.
Since now there are no actions planned for the current (third) cycle the robot waits idly at \texttt{office3} for new requests.
When \rosoclingo\ receives the second request,
the input feeder generates a new input stream update for \oclingo\ initiating the fourth cycle.
The task plan generated by \oclingo\ for satisfying the request is shown in Table \ref{table:example} under plan 3.
Again, assuming the actions are executed without complications the following cycles run analogous to the ones above.
After the sixth cycle the robot delivered the package to \texttt{office4} and enters the idle mode again, awaiting new requests.

%%% Local Variables: 
%%% mode: latex
%%% TeX-master: "paper"
%%% End: 

\section{Conclusion}

Higher level cognitive functions such as reasoning about actions, environment,
goals, or perceptions are crucial in cognitive robotics.
They necessitate knowledge representation and reasoning capacities
for autonomous robots.
% We recently developed a reactive ASP system, called \oclingo,
% which is suitable for handling representation and reasoning capacities in robotics.
We developed a ROS package integrating \oclingo, a reactive 
ASP solver, with the robotics middleware ROS.
The resulting system, called \rosoclingo, 
fulfils the need for high-level knowledge representation 
and reasoning in cognitive robotics by providing a highly expressive and capable 
reasoning framework.
It also makes details of integrating \oclingo\ transparent for the developer.
The developer does not need to deal with parsing answer sets for actions or
transforming other ROS components' results and external events
considering \oclingo\ requirements.
Additionally,
for using \movebase, a widely used ROS package for navigation tasks,
\rosoclingo\ provides an interface layer which also functions as a guideline
for developing interfaces for other ROS components.
Using reactive ASP and \rosoclingo, 
one can control the behaviour of a robot within one framework 
and in a fully declarative way.
This is particularly important compared to Golog based approaches where
the developer should take care of implementation (usually in Prolog)
details of the control knowledge, and the underlying action formalism separately.
We illustrated the usage of \rosoclingo\ via a case-study conducted with 
a ROS-based simulation of a robot delivering mail packages in an office environment 
using \gazebo.

%\bibliographystyle{acmtrans}
%\bibliographystyle{plain}
%\bibliography{local,lit,akku,procs}

\end{document}